\documentclass[10pt,twocolumn,letterpaper]{article}

\usepackage{cvpr}
\usepackage{times}
\usepackage{epsfig}
\usepackage{subcaption}
\usepackage{graphicx}
\usepackage{amsmath}
\usepackage{amssymb}
\usepackage{float}
\usepackage{setspace}
\usepackage{multirow}
\usepackage{mdwlist}

\usepackage[T1]{fontenc}
\usepackage[utf8]{inputenc}
\usepackage{authblk}
\usepackage{array}

\usepackage{color, soul}

\newcommand{\PreserveBackslash}[1]{\let\temp=\\#1\let\\=\temp}
\newcolumntype{C}[1]{>{\PreserveBackslash\centering}p{#1}}
\newcolumntype{R}[1]{>{\PreserveBackslash\raggedleft}p{#1}}
\newcolumntype{L}[1]{>{\PreserveBackslash\raggedright}p{#1}}

\usepackage[pagebackref=true,breaklinks=true,letterpaper=true,colorlinks,bookmarks=false]{hyperref}

\cvprfinalcopy 

\def\cvprPaperID{369} 

\ifcvprfinal\fi
\begin{document}

\sethlcolor{green}

\title{SRN: Side-output Residual Network for Object Symmetry Detection in the Wild}

\author[1,2]{Wei Ke\thanks{This work was supported in part by the CSC, China.}}
\author[2]{Jie Chen}
\author[1]{Jianbin Jiao}
\author[2]{Guoying Zhao}
\author[1]{Qixiang Ye\thanks{Corresponding author}}

\affil[1]{University of Chinese Academy of Sciences\\
Beijing, China
}
\affil[2]{CMVS, University of Oulu, Finland
\authorcr\small \{kewei11\}@mails.ucas.ac.cn, \{jiechen, gyzhao\}@ee.oulu.fi,\{jiaojb, qxye\}@ucas.ac.cn}

\renewcommand\Authands{ and }

\maketitle

\begin{abstract}
\vspace{-0.5em}
   In this paper, we establish a baseline for object symmetry detection in complex backgrounds by presenting a new benchmark and an end-to-end deep learning approach, opening up a promising direction for symmetry detection in the wild. The new benchmark, named Sym-PASCAL, spans challenges including object diversity, multi-objects, part-invisibility, and various complex backgrounds that are far beyond those in existing datasets. The proposed symmetry detection approach, named Side-output Residual Network (SRN), leverages output Residual Units (RUs) to fit the errors between the object symmetry ground-truth and the outputs of RUs. By stacking RUs in a deep-to-shallow manner, SRN exploits the ‘flow’ of errors among multiple scales to ease the problems of fitting complex outputs with limited layers, suppressing the complex backgrounds, and effectively matching object symmetry of different scales. Experimental results validate both the benchmark and its challenging aspects related to real-world images, and the state-of-the-art performance of our symmetry detection approach.  The benchmark and the code for SRN are publicly available at \url{https://github.com/KevinKecc/SRN} .
\end{abstract}

\vspace{-1.5em}
\section{Introduction}
\vspace{-0.5em}

Symmetry is pervasive in visual objects, both in nature creatures like trees and birds, and artificial objects like aircrafts and oil pipes in aerial images. Symmetric parts and their connections {constitute} a powerful part-based decomposition of shapes \cite{17sebastian2004recognition, 31trinh2011skeleton}, providing valuable cue for the task of object recognition. With symmetry constrained, the performance of image segmentation \cite{18teo2015detection}, foreground extraction \cite{19fu2014symmetry}, object proposal \cite{21lee2015learning}, and text-line detection \cite{27zhang2015symmetry} could be significantly improved.

The early symmetry detection, named skeleton extraction, usually involves only binary images \cite{37DBLP:journals/pami/LamLS92,34saha2016survey}. In recent years, symmetry detection tends to process color images \cite{DBLP:conf/cvpr/LiuSZWPLRL13, 23liu2010computational}, but still limited to cropped image patches with little background. This limitation is partially due to the lack of fundamental benchmarks, considering that most existing symmetry detection datasets, \eg, SYMMAX \cite{14tsogkas2012learning}, WH-SYMMAX \cite{15shen2016multiple}, and SK506 \cite{04shen2016object}, lack either object-level annotation or the in-the-wild settings, \ie, multi-objects, part-invisibility, and various complex backgrounds.

\begin{figure}[t]
\begin{center}
   \includegraphics[width=0.99\linewidth]{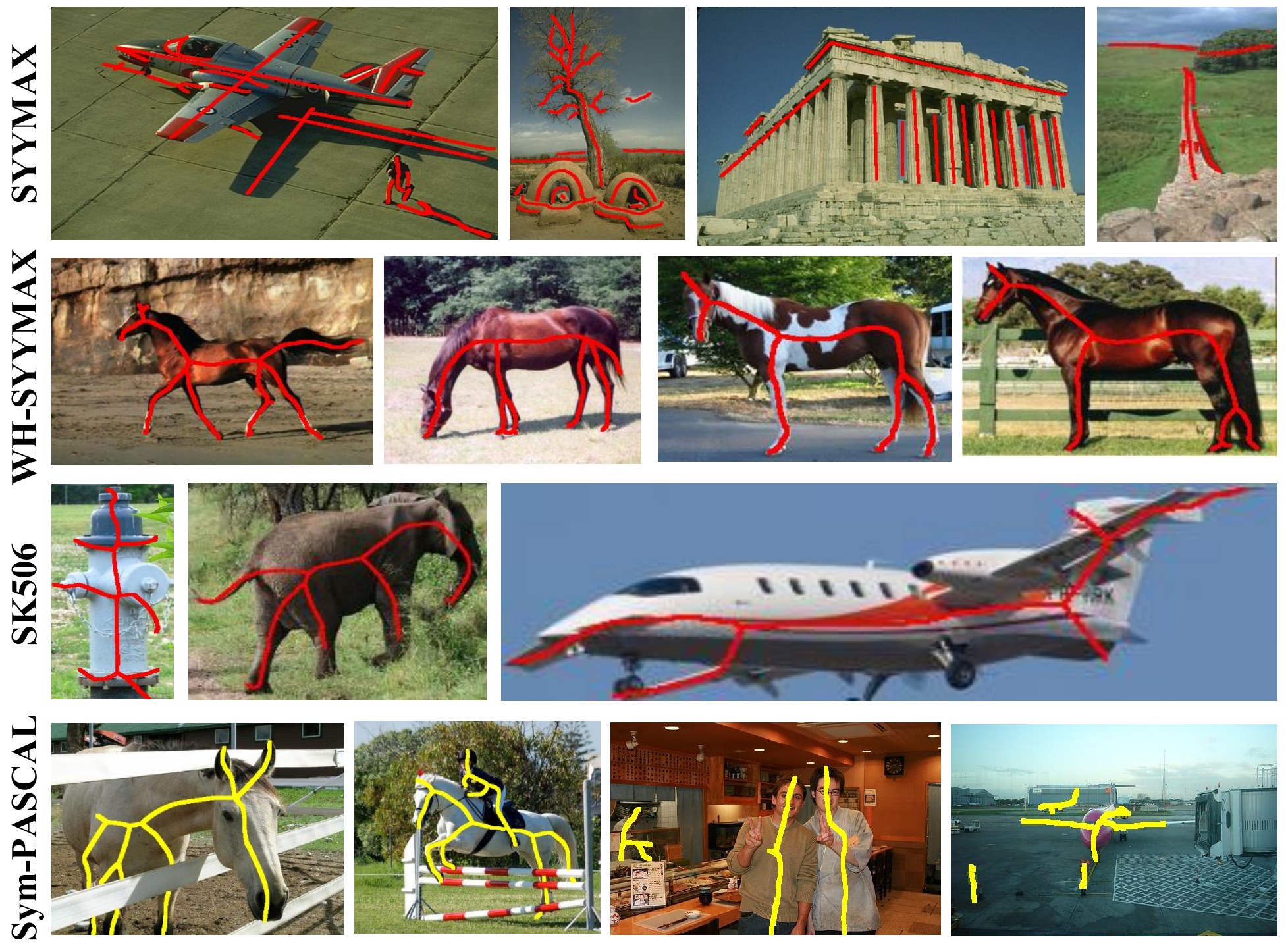}
\end{center}
\vspace{-1.5em}
   \caption{We propose a new benchmark, named Sym-PASCAL, for object symmetry detection in the wild. Compared with SYMMAX \cite{14tsogkas2012learning}, WH-SYMMAX \cite{15shen2016multiple}, and SK506 \cite{04shen2016object}, our Sym-PASCAL spans challenges including object diversity, multi-objects, part-invisibility and various complex backgrounds. (Best viewed in color) }
   \vspace{-1.5em}
\label{figure1}
\end{figure}

In this paper, we present a new challenging benchmark with complex backgrounds, and an end-to-end deep symmetry detection approach that processes in-the-wild images, and target at opening up a promising direction for practical applications of symmetry. The new benchmark, named Sym-PASCAL, is composed of 1453 natural images with 1742 objects derived from the PASCAL-VOC-2011 \cite{10pascalvoc2011} segmentation dataset. Such a benchmark is more close to practical applications with challenges far beyond those in existing datasets: (1) \emph{diversity of objects}: multi-class objects with different illuminations and viewpoints; (2) \emph{multi-object co-occurrence}: multiple objects exist in a single  image; (3) \emph{part-invisibility}: objects are partially occluded; and (4) \emph{complex backgrounds}: the scenes where object located could be contextually cluttered.

For the in-the-wild symmetry detection problem, we explore the deep Side-output Residual Network (SRN) that directly outputs response image about object symmetry. SRN roots in the Holistically-nested Edge Detection (HED) network \cite{03xie2015holistically} but updates it by stacking multiple Residual Units (RUs) on the side-outputs. The Residual Unit (RU) is designed to fit the error between the object symmetry ground-truth and the outputs of RUs, which is computationally easier as it pursuits the minimization of residuals among scales rather than only struggles to combine multi-scale features to fit the object symmetry ground-truth. The RU we defined not only significantly improves the performance of SRN, but also solves the learning convergence problem left by the baseline HED method. By stacking multiple RUs in a deep-to-shallow manner, the receptive fields of stacked RUs could adaptively match the scale of symmetry. The contributions of this paper include:
\vspace{-1em}
\begin{itemize*}
\item {A new object symmetry benchmark} that spans challenges of diversity, multi-objects, part-invisibility, and various complex backgrounds, promoting the symmetry detection research to in-the-wild scenes.
\item {A Side-output Residual Network that can effectively fit the errors between ground-truth and the outputs of the stacked RUs, enforcing the modeling capability to symmetry in complex backgrounds, achieving state-of-the-art symmetry detection performance in the wild.}
\vspace{-1.5em}
\end{itemize*}

\vspace{-1.0em}
\section{Related Works}
\vspace{-0.5em}

For the applicability and beauty, symmetry has attracted much attention in the past decade. The targets of symmetry detection evolute from binary images to color object images, while the symmetry detection approaches update from hand-crafted to learning based.

\textbf{Benchmarks:} In the early research, symmetry extraction algorithms are qualitatively evaluated on quite limited binary shapes \cite{37DBLP:journals/pami/LamLS92}. Such shapes are selected from the MPEG-7 Shape-1 dataset for subjective observation \cite{11bai2007skeleton}. Later, Liu \etal \cite{DBLP:conf/cvpr/LiuSZWPLRL13} use very a few real-world images to perform symmetry detection competitions. To be honest, SYMMAX \cite{14tsogkas2012learning} could be regarded as an authentic benchmark that contains hundreds of training/testing images with local symmetry annotation. But the local reflection symmetry it defined mainly focuses on low-level image {edges and contours}, missing the high-level concept of objects. WH-SYMMAX \cite{15shen2016multiple} and SK506 \cite{04shen2016object} are recently proposed benchmarks with annotation of object skeletons. Nevertheless, WH-SYMMAX is simply composed of side-view horses while SK506 consists objects with little background. Neither of them involves multiple objects in complex backgrounds, leaving a plenty of room for developing new object symmetry benchmarks.

\textbf{Methods:} Early symmetry detection methods, also named skeleton extraction \cite{37DBLP:journals/pami/LamLS92,34saha2016survey}, are mainly developed for the binary images by leveraging morphological image operations. When processing color images, they usually need a contour extraction or an image segmentation step as pre-processing. Considering that segmentation of in-the-wild images remains a research problem, the integration of color image segmentation and symmetry detection not only increases the complexity but also accumulates the errors.

Researchers have tried to extract symmetry in color images based on multi-scale super-pixels. One hypothesis is that the object symmetry axes are the subsets of lines connecting the center points of super-pixels \cite{29levinshtein2009multiscale}. Such line subsets are explored from the super-pixels using a sequence of deformable disc models extracting the symmetry pathes \cite{30sie2013detecting}. Their consistence and smoothness are enforced with spatial filters, e.g., a particle filter, which link local skeleton segments into continuous curves \cite{36widynski2014local}. Due to the lack of object prior and the learning module, however, these methods are still limited to handle the images with simple backgrounds.

More effective symmetry detection approaches root in powerful learning methods. On the SYMMAX benchmark, the Multiple Instance Learning (MIL) \cite{14tsogkas2012learning} is used to train a curve symmetry detector with multi-scale and multi-orientation features. To capture diversity of symmetry patterns, Teo \etal \cite{18teo2015detection} employ the Structured Random Forest (SRF) and Shen \etal \cite{15shen2016multiple} use subspace MIL with the same feature. Nevertheless, as the pixel-wise hand-craft feature is computationally expensive and representation limited, these methods are intractable to detect object symmetry in complex backgrounds.

Most recently, a deep learning approach, Fusing Scale-associated Deep Side-outputs (FSDS) \cite{04shen2016object}, is shown to be capable of learning unprecedentedly effective object skeleton representations on WH-SYMMAX \cite{15shen2016multiple} and SK506 \cite{04shen2016object}. FSDS takes the architecture of HED \cite{03xie2015holistically} and supervises {its side-outputs} with scale-associated ground-truth. Despite of its state-of-the-art performance, it needs the intensive annotations of the scales for each skeleton point, which means that it uses much more human effort than other approaches when preparing the training data. Compared with FSDS, our proposed SRN can adaptively match the scales of symmetry, without using scale-level annotation.

\vspace{-1em}
\section{The Sym-PASCAL Benchmark}
\vspace{-0.5em}

Symmetry annotation involves pixel-level fine details, and is time consuming. We thus leverage the semantic segmentation ground-truth and a skeleton generation algorithm to aid the annotation of symmetry \cite{38DBLP:journals/pr/ShenBHWL11}.

\begin{figure}[t]
\begin{subfigure}{0.153\textwidth}
\centering
\includegraphics[height=10em]{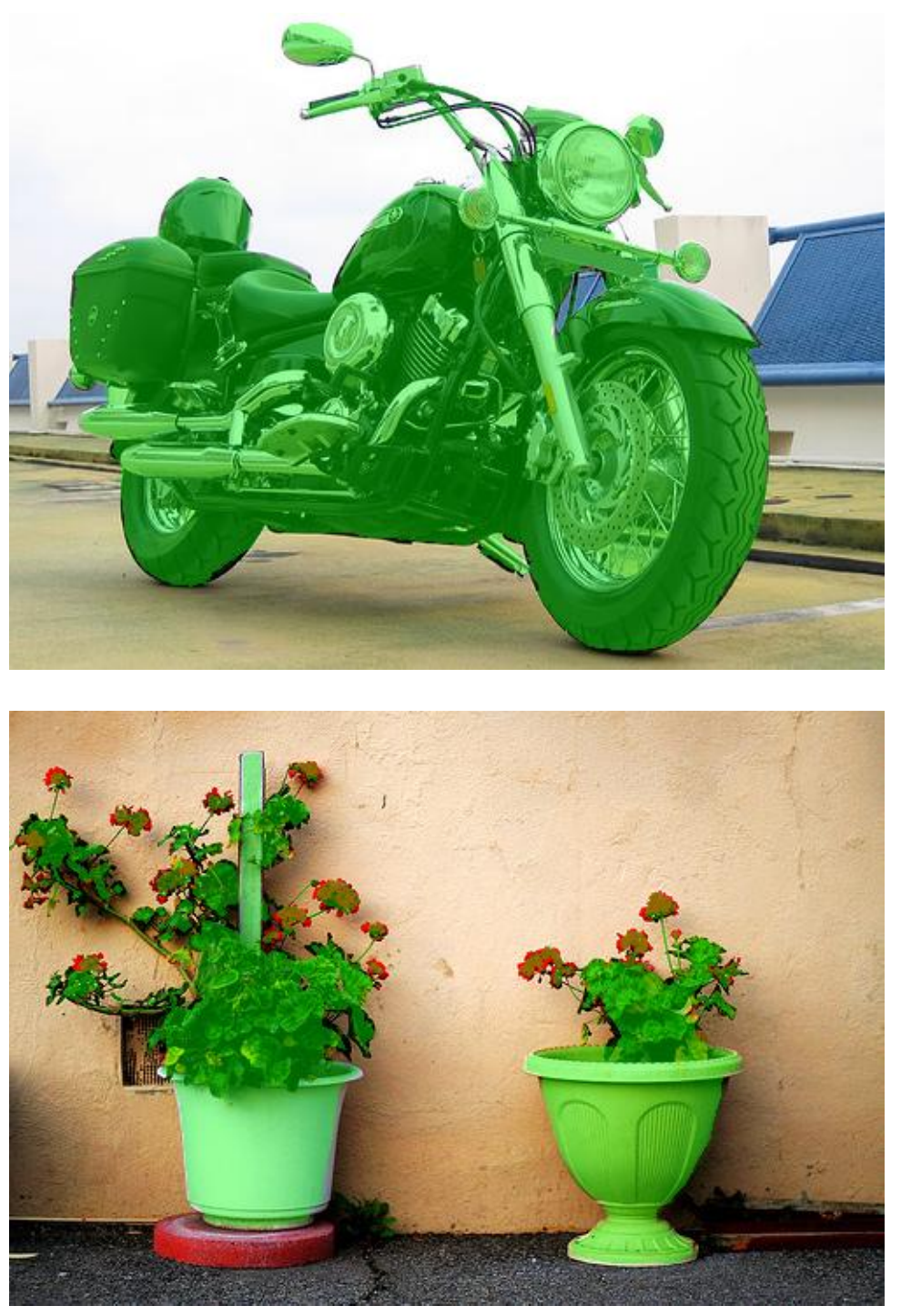}
\caption{Unavailable} \label{figure3a}
\end{subfigure}
\begin{subfigure}{0.163\textwidth}
\centering
\includegraphics[height=10em]{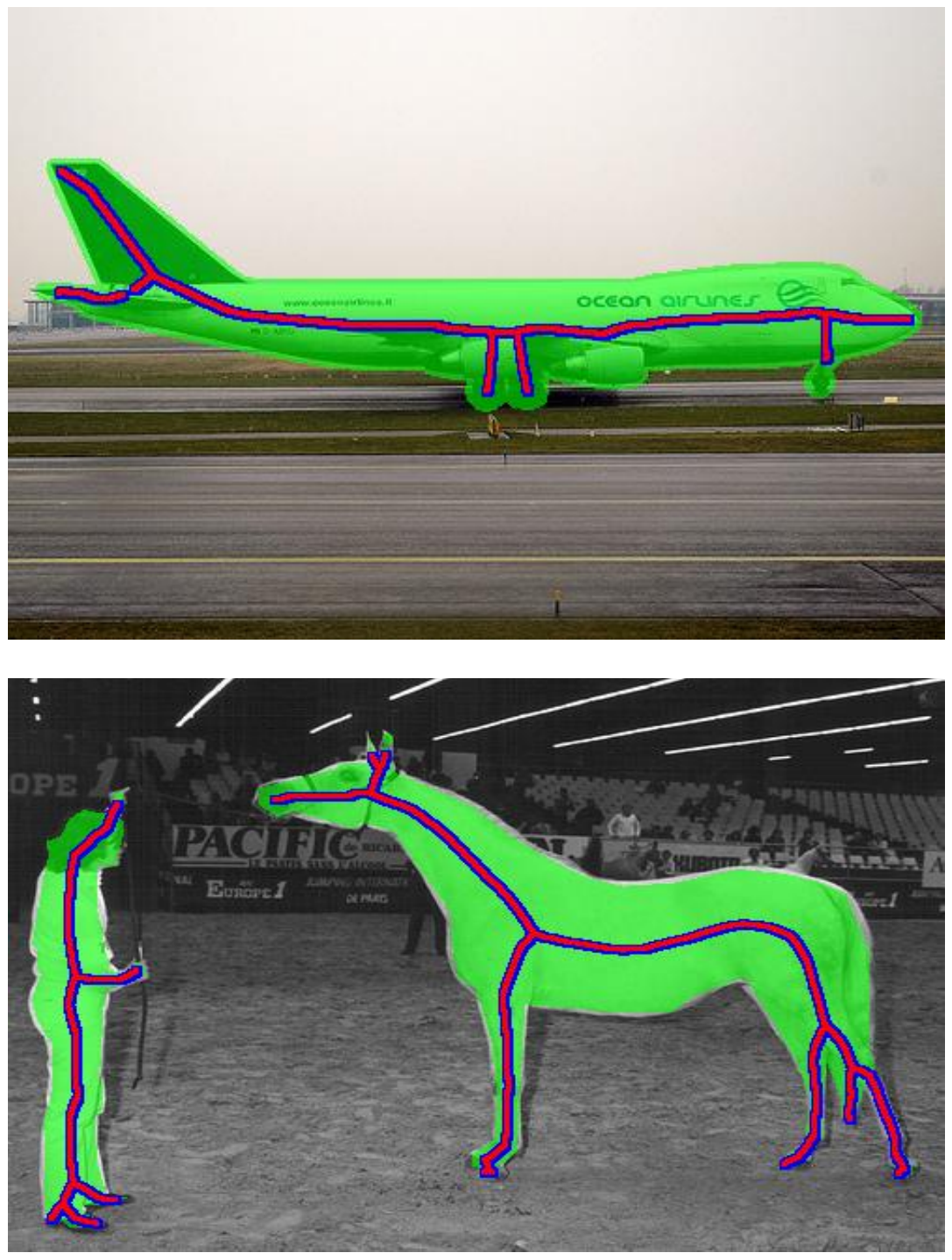}
\caption{Available (easy)} \label{figure3b}
\end{subfigure}
\begin{subfigure}{0.153\textwidth}
\centering
\includegraphics[height=10em]{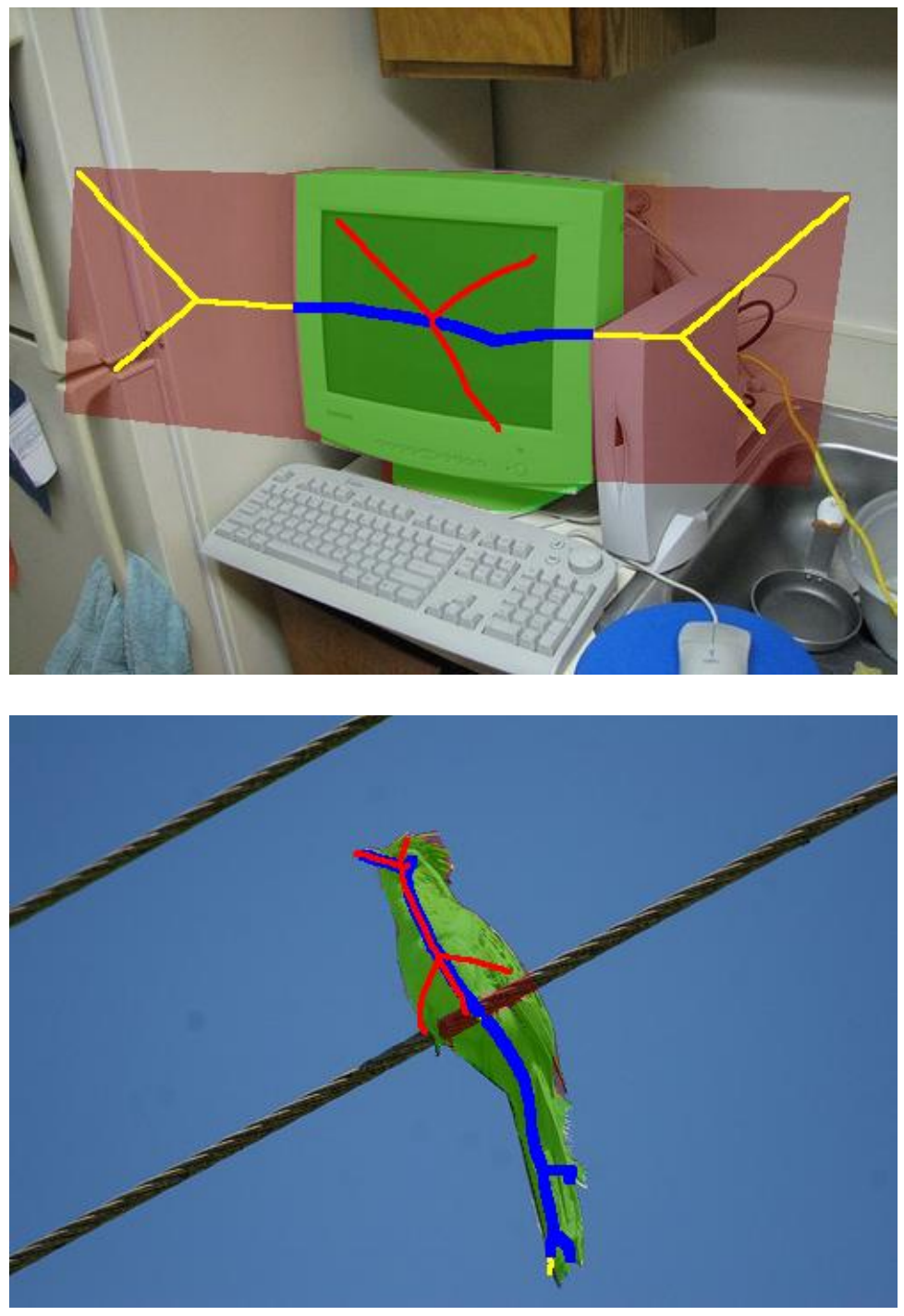}
\caption{Available (hard)} \label{figure3c}
\end{subfigure}
\vspace{-0.8em}
   \caption{Object symmetry annotation. The green masks are annotated semantic segmentation ground-truth. The brown masks are extended from the segmentation. The red lines are the skeletons of semantic segmentation masks. The yellow and blue lines are the skeletons corresponding to the extended masks. The blue lines are the object symmetry ground-truth. (Best viewed in color)}
\vspace{-1.8em}
\label{figure3}
\end{figure}

\subsection{Categorization and Annotation}
\vspace{-0.2em}

Sym-PASCAL is derived from the PASCAL-VOC-2011 segmentation dataset \cite{10pascalvoc2011} which contains 1112 training images and 1111 testing images from 20 object classes including: person, bird, cat, cow, dog, horse, sheep, aero plane, bicycle, boat, bus, car, motorbike, train, bottle, chair, dining table, potted plant, sofa, and tv/monitor.

We categorize the 20 classes of objects into symmetry-available and symmetry-unavailable, Fig.\ \ref{figure3}. The objects that contain lots of discontinuous parts in the segmentation masks are symmetry-unavailable, specifically potted plant, dining table, motorbike, bicycle, chair and sofa, are not selected, Fig.\ \ref{figure3a}. The other 14 object classes are symmetry-available. Some of objects are slender and thus easy to annotate, Fig.\ \ref{figure3b}, and others with small length-width ratio or occlusion are difficult to annotate, Fig.\ \ref{figure3c}. In total, 648/787 images are selected and annotated from the PASCAL-VOC-2011 training and testing sets. Among these images, 31.3\% are with multi-object and 45.6\% are with part-invisibility.

{For the images where object symmetry is obvious, \ie, objects are composed of slender parts that are easy to annotate, we directly extract symmetry on the object segmentation masks using a skeleton extraction algorithm \cite{38DBLP:journals/pr/ShenBHWL11}, Fig.\ \ref{figure3b}. For such objects, the object symmetry (marked with blue curves) and their skeleton (marked with red curves) are consistent. For the images where object symmetry is not obvious, we manually extend the semantic segmentation masks and annotate symmetry on them, Fig.\ \ref{figure3c}. For wide object as shown on the top of Fig.\ \ref{figure3c}, we extend the mask along the direction of the long axis of the object and choose the long axis as ground-truth. For occluded objects as shown at the bottom of Fig.\ \ref{figure3c}, we need to manually fill the missed parts of segmentation masks. For the pictures that contain partial objects, we empirically imagine the occluded parts to extend the segmentation masks. With these processing above, the skeleton extraction algorithm \cite{38DBLP:journals/pr/ShenBHWL11} is used to extract symmetry on the object segmentation masks. The object symmetry ground-truth is set as the skeleton points within the segmentation masks, shown as the blue curves in Fig.\ \ref{figure3c}.}

\begin{figure}[b]
\vspace{-0.5em}
\begin{subfigure}{0.24\textwidth}
\centering
\includegraphics[width=\linewidth, height=6.5em]{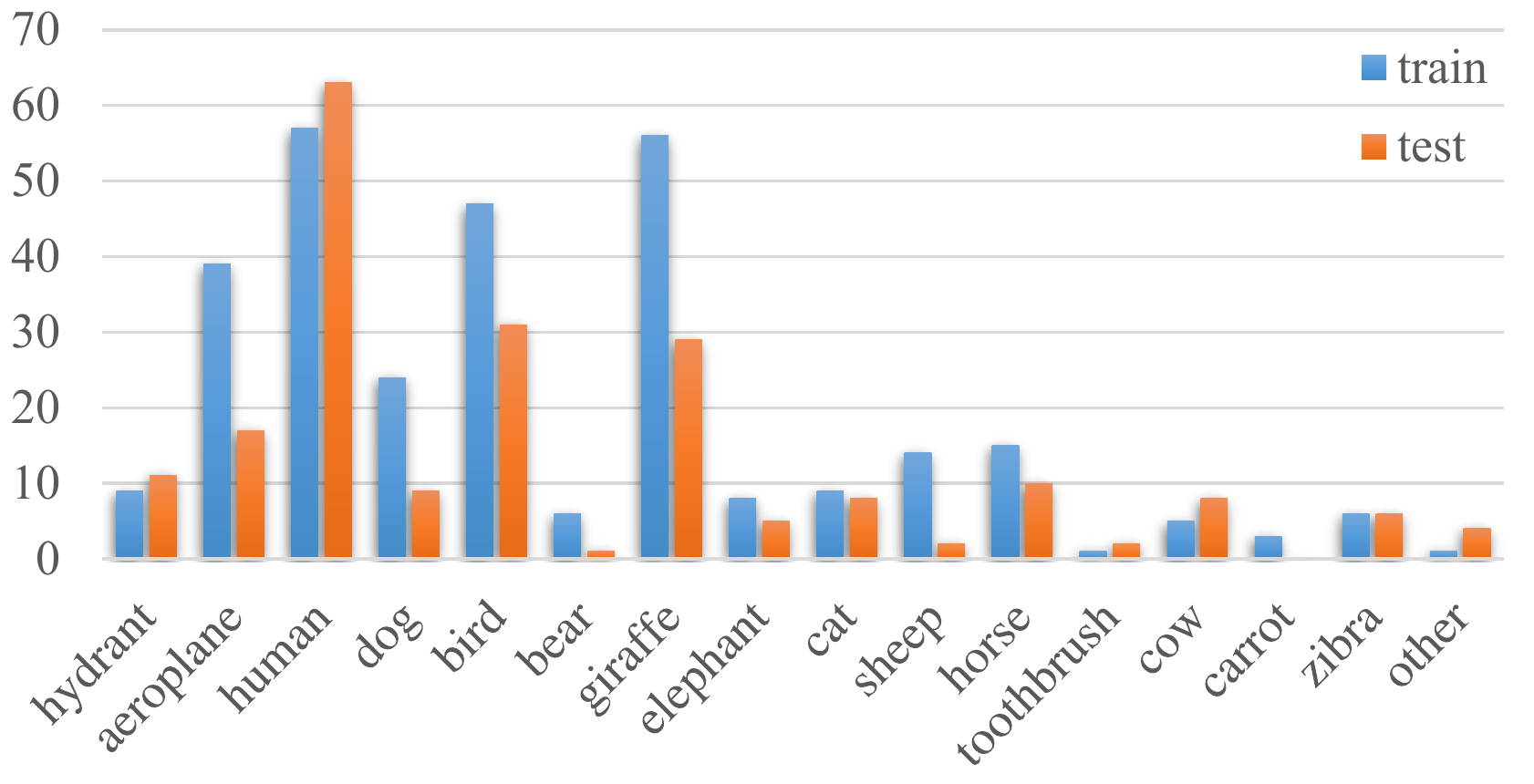}
\caption{SK506} \label{figure4a}
\end{subfigure}
\begin{subfigure}{0.23\textwidth}
\centering
\includegraphics[width=\linewidth, height=6.5em]{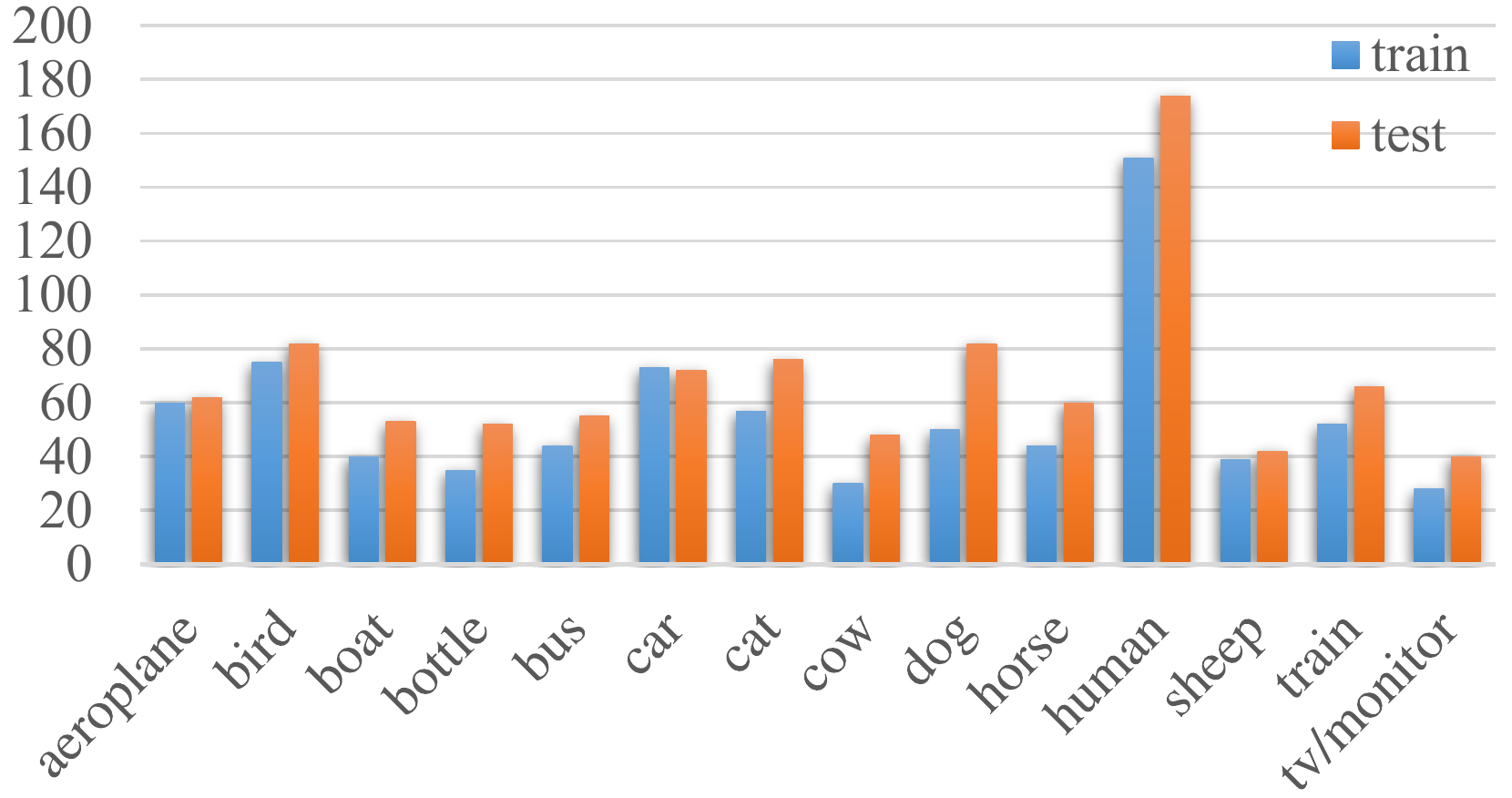}
\caption{Sym-PASCAL} \label{figure4b}
\end{subfigure}
\vspace{-0.5em}
\caption{Object-class distributions of the SK506 and Sym-PASCAL datasets.} \label{figure4}
\vspace{-0em}
\end{figure}

\begin{table} [b]
\begin{center}
\small
\begin{tabular}{L{0.15\linewidth}|C{0.17\linewidth}C{0.14\linewidth}C{0.11\linewidth}C{0.17\linewidth}}
\hline
\multirow{2}{*}{}           & \multirow{2}{*}{SYMMAX} & WH-      & \multirow{2}{*}{SK506} & Sym-     \\
                            &                         & SYMMAX   &                        & PASCAL   \\
\hline
Data                            & local                   & object   & object                 & object   \\
type                            & symmetry                & skeleton & skeleton               & symmetry \\
\hline
Image                           & in-the-wild             & simple   & simple                 & in-the-wild  \\
type                            & image                   & image    & image                  & image    \\
\hline
\#object                    & --                      & 1        & 16                     & 14       \\
\hline
\#training                 & 200                     & 228      & 300                    & 648      \\
\hline
\#testing                   & 100                     & 100      & 206                    & 787      \\
\hline
\end{tabular}
\end{center}
\vspace{-1em}
\caption{Comparison of four symmetry detection datasets.}
\label{Tab-1}
\end{table}

\begin{figure*}[t]
\vspace{-0.5em}
\begin{subfigure}{0.49\textwidth}
\centering
\includegraphics[width=0.95\linewidth]{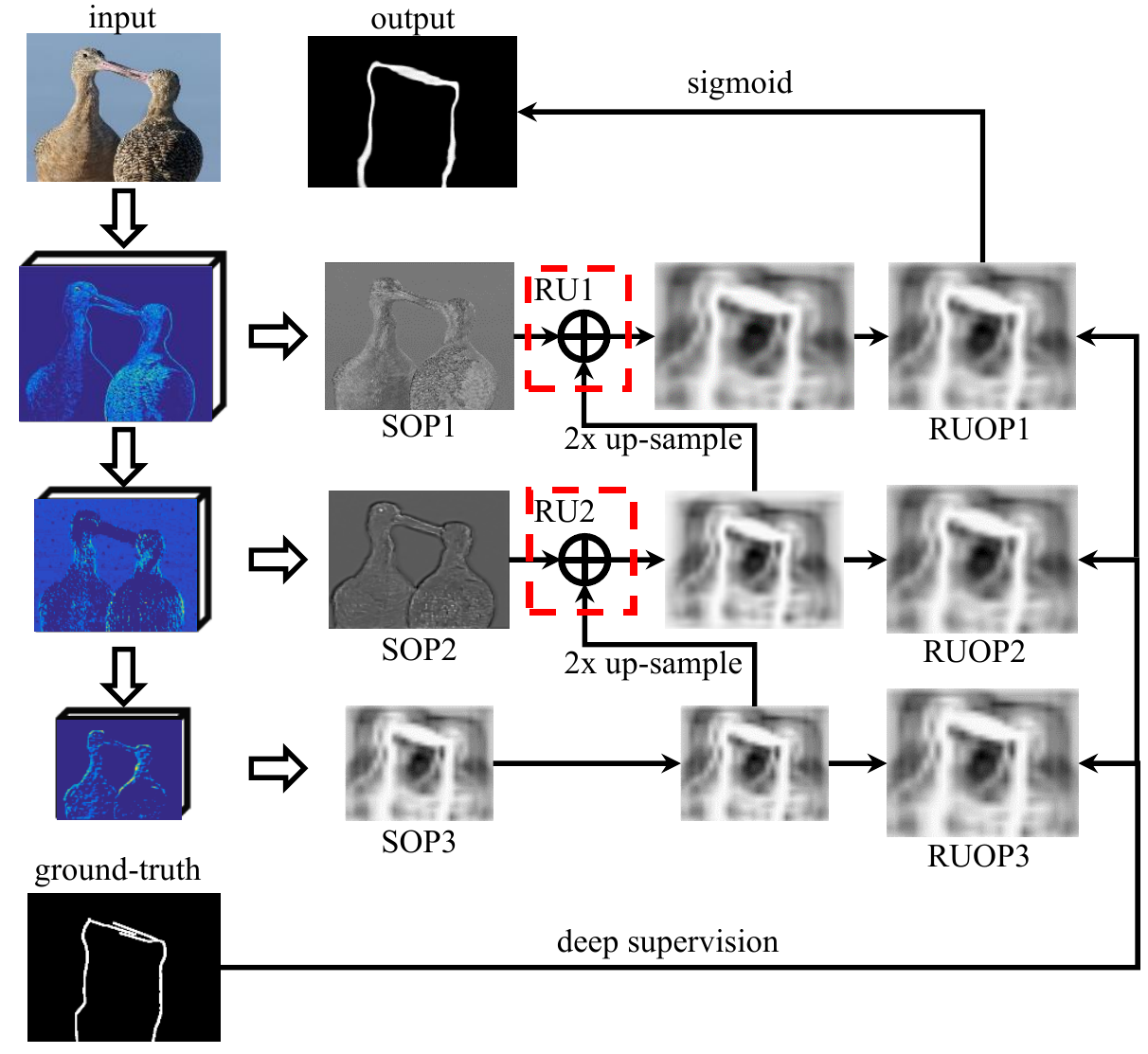}
\caption{Deep-to-shallow} \label{figure6a}
\end{subfigure}
\hspace*{\fill} 
\begin{subfigure}{0.49\textwidth}
\centering
\includegraphics[width=0.95\linewidth]{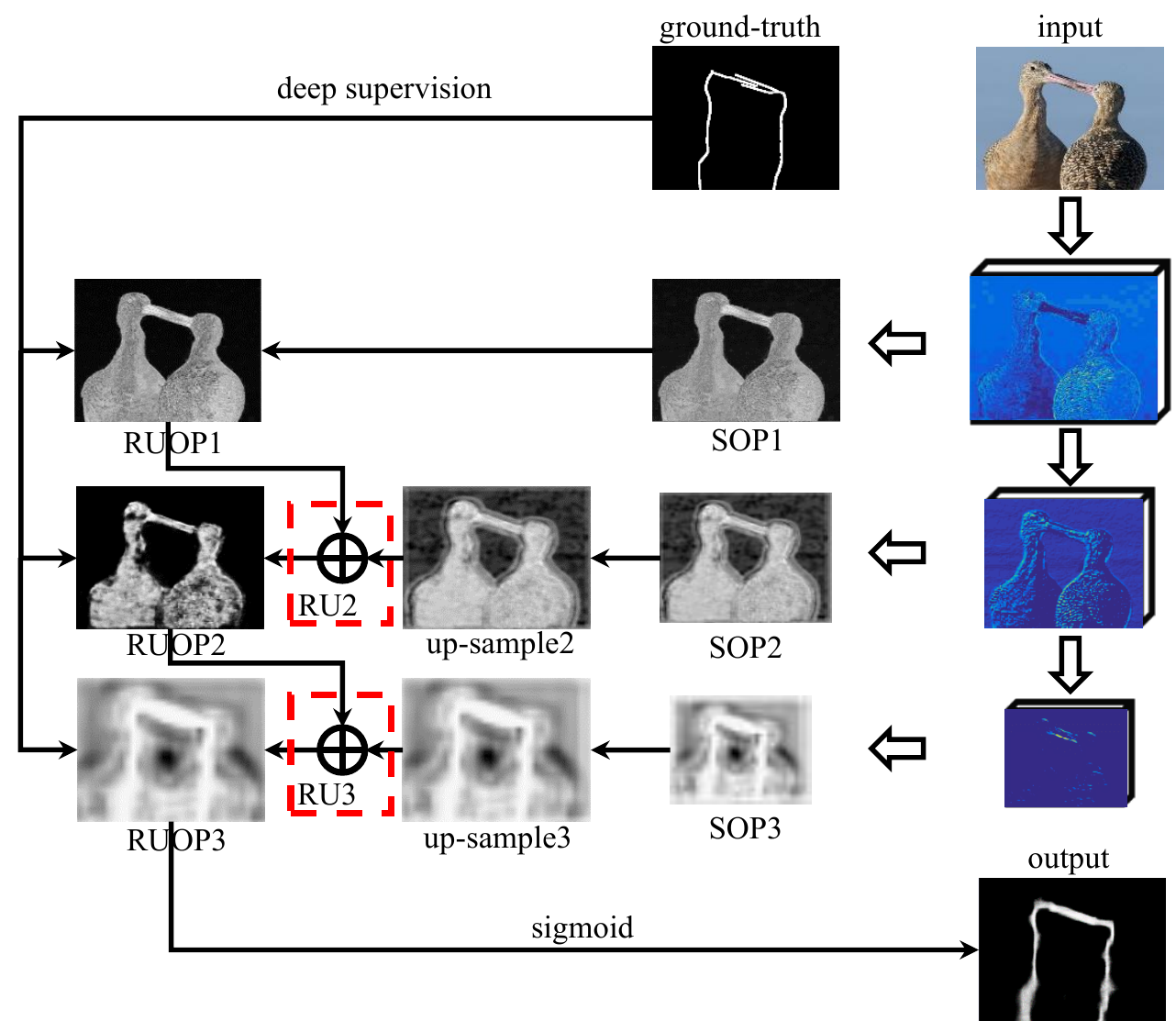}
\caption{Shallow-to-deep} \label{figure6b}
\end{subfigure}
\vspace{-1em}
\caption{The architectures of proposed Side-Output Residual Network (SRN) by stacking Residual Units (RUs) in (a) deep-to-shallow and (b) shallow-to-deep strategies. The RUs are marked with dashed boxes. With the deep supervision both on the input and output of RU, the residual between the ground-truth and output of RU (RUOP) is computed hierarchically. Along the stacking orientation, the residual decreases so that the RUOP is closer to ground-truth.} \label{figure6}
\vspace{-1em}
\end{figure*}

\vspace{-0.5em}
\subsection{Discussion}
\vspace{-0.5em}

In what follows, we compare the proposed benchmark with three other representative ones, SYMMAX \cite{14tsogkas2012learning}, WH-SYMMAX \cite{15shen2016multiple}, and SK506 \cite{04shen2016object}.

SYMMAX is derived from BSDS300 \cite{16arbelaez2011contour}, which contains 200/100 training and testing images. It’s annotated with local reflection symmetry on both foreground and background. Considering that most computer vision tasks focus on the foreground, it’s more meaningful to use object symmetry instead of the symmetry about the whole image. WH-SYMMAX is developed for object skeletons, but it is made up of only cropped horse images, which are not comprehensive for general object symmetry. SK506 involves skeletons about 16 classes of objects. Nevertheless, their backgrounds are too simple to represent in-the-wild images.

As shown in Tab.\ \ref{Tab-1}. the proposed benchmark involves more training and testing images. Particularly, these images involve complex backgrounds, multiple objects and/or occlusions. It is developed for end-to-end object symmetry in-the-wild, providing the protocol to evaluate whether or not an algorithm can detect symmetry without using additional object detectors. In Sym-PASCAL, the images for each class are more balanced than other datasets, Fig.\ \ref{figure4b}, except that the number of human objects is larger than others. In contrast, in SK506 the objects from different classes have more unbalance, Fig.\ \ref{figure4a}.

\section{Side-output Residual Network}

The proposed Side-output Residual Network (SRN) roots in the well-designed output Residual Unit (RU) and a deep-to-shallow learning strategy. Given the symmetry ground-truth, the SRN is learned in an end-to-end manner.

\subsection{Output Residual Unit}

\begin{figure}[t]
\vspace{-1em}
\begin{center}
   \includegraphics[width=0.45\linewidth]{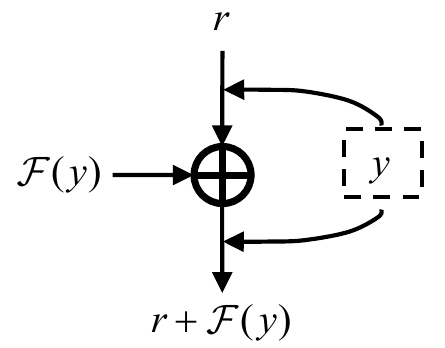}
\end{center}
\vspace{-2em}
   \caption{The output Residual Unit (RU). By supervision both on the input and output of RU, the additional mapping ${\cal F}(y)$ estimates the residual of $y$.}
   \vspace{-1.5em}
\label{figure5}
\end{figure}

Given training images, the end-to-end symmetry learning pursuits deep network parameters that best fit the symmetry ground-truth. Such a learning objective is different from that of learning a classification network \cite{07he2015deep}. The RU defined for output, Fig.\ \ref{figure5}, is essentially different from that in the residual network defined for features \cite{07he2015deep}. With the deep supervision both on the input and output of RUs, the residual of the ground-truth is computed. Formally, denoting the input of RU as $r$ and the additional mapping as ${\cal F}(y)$, the deep supervision is written as:
\begin{equation}
\left\{ {\begin{array}{*{20}{c}}
{r \approx y}\\
{r + {\cal F}(y) \approx y}
\end{array}} \right.{\rm{ ,}}
\label{Eq1}
\end{equation}
where $r$ and $r + {\cal F}(y)$ are the input and output of the RU, respectively. ${\cal F}(y)$ is regarded as the residual estimation of $y$. RUs provide shortcut connections between the ground-truth and outputs from different scales, which implies a functional module for the `flow' of errors among different scales, and thus make it easier to fit complex outputs with higher adaptivity. To the extreme, if an input $r$ is optimal, it would be easier to push the residual to zero than to fit the additional mapping ${\cal F}(y)$.

\subsection{Network Architectures}

By stacking the RUs defined, we implement a kind of new side-output deep network, named Side-output Residual Network (SRN), which incorporates the advantages of both the scale adaptability and residual learning. For SRN, the input of the first RU can be chosen as the shallowest side-output or deepest side-output, which derives two versions of SRN, Fig.\ \ref{figure6}. In what follows, the RU is numbered as the side-output (SOP) index, and the output of the $i{\rm{ - th}}$ RU is denoted as ${\rm{RUOP}}i$, for short.

\textbf{Deep-to-shallow.} In this SRN architecture, RUs are stacked from deep to shallow, Fig.\ \ref{figure6a}. Assume that ${s_i}$ is the $i$-th side-output, and ${r_{i + 1}}, {r_i}$ are the input and output of $i$-th RU respectively . For the first stacked RU2, the input is set as the deepest SOP3, \ie, ${r_3} = {s_3}$. And SOP2 is used to learn the residual between RUOP3 and the ground-truth, which updates RUOP3 to RUOP2. The RUs are stacked in order until the shallowest side-output, in other words, the inputs of which are set as the output of the former one. Sigmoid is used as classifier on the output of the last stacked RU to generate the final output image.

\begin{figure}[t]
\begin{subfigure}{0.20\textwidth}
\centering
\includegraphics[width=0.9\linewidth, height=9em]{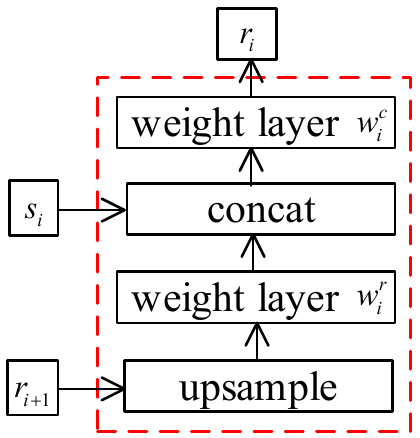}
\caption{Deep-to-shallow} \label{figure7a}
\end{subfigure}
\hspace*{\fill} 
\begin{subfigure}{0.20\textwidth}
\centering
\includegraphics[width=\linewidth,height=9em]{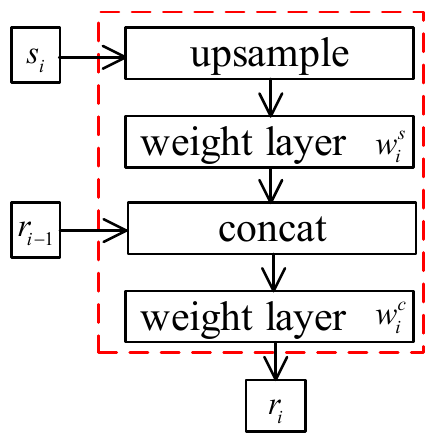}
\caption{Shallow-to-deep} \label{figure7b}
\end{subfigure}
\vspace{-1em}
\caption{The implementation of the $i$-th RU.} \label{figure7}
\vspace{-1em}
\end{figure}

The implementation of RU in the deep-to-shallow architecture is shown in Fig.\ \ref{figure7a}. It’s noting that the output size of RU in this architecture is same as the side-output rather than the input image. Therefore, a Gaussian deconvolution layer is introduced to the output of RU. As the up-sampling is non-linear transformation, a weight layer is stacked to improve the scale adaptability. Instead of adding up-sampled ${r_{i + 1}}$ and $s_i$ directly, a $1 \times 1$ convolutional layer is utilized to generate ${r_i}$. The RU is formulated,
\begin{equation}
{r_i} = {w_i}^c({s_i} + w_i^r{r_{i + 1}}),
\label{Eq2}
\end{equation}
where ${w_i}^c,{w_i}^r$ are the convolutional weights of concatenation layer and the up-sampled ${r_{i + 1}}$. With Eqs. (\ref{Eq1}) and (\ref{Eq2}), the output residual ${{\cal F}_i}(y)$ is computed,
\begin{equation}
{{\cal F}_i}(y) = w_i^c \cdot {s_i} + (w_i^rw_i^c - 1){r_{i + 1}}.
\label{Eq3}
\end{equation}
When $w_i^r \cdot w_i^c$ approximates 1.0, the residual is related to only the side-output. To the extreme, along the stacking orientation of RUs, the residual ${\cal F}(y)$ approximates 0.0.

As we know, the deep layers of CNNs contain features that ignore the image details but capture high-level representations. Therefore, a deep layer SOP3 is expected to be closer to the optimal training solution. RU2 pushes the residual to zero and the response map RUOP2 is similar with the response map RUOP3. In the deep-to-shallow architecture, the deepest side-output is used as a good initialization for the ground-truth, therefore, the deep-to-shallow architecture contributes better results than the shallow-to-deep one, as shown in Sec.\ \ref{SRN-setting}.

\textbf{Shallow-to-deep.} The architecture is shown in Fig.\ \ref{figure6b} and the RU in Fig.\ \ref{figure7b}. The side-outputs are up-sampled by the Gaussian deconvolution layer so that their size is consistent with the input image. Similar with Eq. (\ref{Eq3}), the residual is computed,
\begin{equation}
{{\cal F}_i}(y) = w_i^sw_i^c \cdot {s_i} + (w_i^c - 1){r_{i + 1}},
\end{equation}
where ${w_i}^s$ is weight parameter of the up-sampled ${s_i}$.
	Fig.\ \ref{figure6b} indicates that the shallowest RUOP1 has lots of false positive pixels compare to ground-truth as SOP1 represents local structure of the input image. Along the stacking orientation, the RU3 reduces the residual so that the outputs of RU3, \ie, RUOP3, are closer to ground-truth compared to RUOP2.

\subsection{Learning}
Given the object symmetry detection training dataset $S = \{ ({X_n},{Y_n})\} _{n = 1}^N$  with $N$ training pairs, where ${X_n} = \{ x_j^{(n)},j = 1, \cdots ,T\}$ and ${Y_n} = \{ y_j^{(n)},j = 1, \cdots ,T\} $ are the input image and the ground-truth binary image with $T$ pixels, respectively. $y_j^{(n)} = 1$ denotes the symmetry pixel and $y_j^{(n)} = 0$ denotes non-symmetry pixel. We subsequently drop the subscript $n$ for notational simplicity, since we consider each image independently. We denote ${\bf{W}}$ as the parameters of the base network. Supposing the network has $M$ side-outputs, the $M$-th side-output is set as the basic output and $M-1$ RUs are used. We use the architecture of Fig.\ \ref{figure6a} as example, in which $M=3$ and RUOP3 is the basic output. Fig.\ \ref{figure6b} has similar formulation. For the basic output, the loss is computed,
\begin{equation}
\begin{array}{l}
{{\cal L}_b}({\bf{W}},{w_b}) =  - \beta \sum\limits_{j \in {Y_ + }} {\log \Pr ({y_j} = 1|X;{\bf{W}},{w_b})} \\
{\rm{               }} - (1 - \beta )\sum\limits_{j \in {Y_ - }} {\log \Pr ({y_j} = 0|X;{\bf{W}},{w_b})} ,
\end{array}
\end{equation}
where ${w_b}$ is the classifier parameter for the basic output. $Y_+$ and $Y_-$ respectively denote the symmetry and non-symmetry ground-truth label sets. The loss weight $\beta  = {{|{Y_ + }|} \mathord{\left/
 {\vphantom {{|{Y_ + }|} {|Y|}}} \right.
 \kern-\nulldelimiterspace} {|Y|}}$, and $|{Y_ + }|$ and $|{Y_ - }|$ denote the symmetry and non-symmetry pixel number, respectively. $\Pr ({y_j} = 1|X;{\bf{W}},{w_b}) \in \left[ {0,1} \right]$ is the sigmoid prediction of the basic output that measures how likely the point to be on the symmetry axis. For the $i$-th RU, $i = M-1, \cdots ,1$, the loss is computed,
\begin{equation}
\begin{array}{l}
{{\cal L}_i}({\bf{W}},{\theta _i},{w_i}) =  - \beta \sum\limits_{j \in {Y_ + }} {\log \Pr ({y_j} = 1|X;{\bf{W}},{\theta _i},{w_i})} \\
{\rm{               }} - (1 - \beta )\sum\limits_{j \in {Y_ - }} {\log \Pr ({y_j} = 0|X;{\bf{W}},{\theta _i},{w_i})}
\end{array}
\end{equation}
where ${\theta _i} = (w_i^c,w_i^s)$ is the convolutional parameter of the concatenation layers and side-output layers after the $i$-th RU. ${w_i}$ is the classifier parameter for the output of $i$-th RU. The loss function for all the stacked RUs is obtained by
\begin{equation}
{\cal L}({\bf{W}},\theta ,w) = {\alpha _M}{{\cal L}_b}({\bf{W}},{w_b}) + \sum\limits_{i = {M-1}}^{1} {{\alpha _i}{{\cal L}_b}({\bf{W}},{\theta _i},{w_i})} .
\end{equation}
Finally, we obtain the optimal parameters,
\begin{equation}
{({\bf{W}},\theta ,w)^*} = \arg \min {\cal L}({\bf{W}},\theta ,w).
\end{equation}

In the testing phase, giving an image $X$, a symmetry prediction map is output by the last stacked RU,
\begin{equation}
\hat Y = \Pr ({y_j} = 1|X;{{\bf{W}}^*},{\theta ^*},{w^*}).
\end{equation}

\subsection{Difference to Other Networks}
\vspace{-0.3em}
The proposed SRN has significant difference with other end-to-end deep learning implementations, \ie, HED \cite{03xie2015holistically}, FSDS \cite{04shen2016object}, and Laplacian Reconstruction \cite{05ghiasi2016laplacian}. In HED, the deep supervision is applied on side-outputs directly, while in SRN the deep supervision is applied on the outputs of RUs. According to (\ref{Eq2}), each RU contains the information of two side-outputs at least, endowing SRN with the capability to smoothly model the multi-scale symmetry across deep layers. FSDS is an improvement of HED that specifies scales for side-outputs, which requires additional annotation for each scale. In contrast, SRN models the scale information with RUs, without any multi-scale annotations. SRN takes the idea of Laplacian reconstruction that uses a mask to indicate the reconstruction residual for segmentation. The difference lies in that SRN pursuits scale adaptability while the Laplacian reconstruction focuses on multi-scale error minimization.

\section{Experimental results}

The proposed SRN is first evaluated and compared on the proposed Sym-PASCAL benchmark. It is then evaluated and compared with the state-of-the-art deep learning approaches on other popular datasets including SYMMAX \cite{14tsogkas2012learning}, WH-SYMMAX \cite{15shen2016multiple}, and SK506 \cite{04shen2016object}.
\vspace{-0.2em}
\subsection{Experimental Setup}
\vspace{-0.3em}
\textbf{Implementation details.} The SRN is implemented following the parameter setting of HED \cite{03xie2015holistically}, by fine-tuning the pre-trained 16-layer VGG net \cite{01simonyan2014very}. The hyper-parameters of SRN include: mini-batch size (1), learning rate (1e-8 for in-the-wild image datasets and 1e-6 for simple image datasets), loss-weight  for each RU output (1), momentum (0.9), and initialization of the nested filters (0), weight decay (0.002), and maximum number of training iterations (18,000). In the testing phase, a non-maximal suppression (NMS) algorithm \cite{09dollar2015fast} is applied on the output map to obtain object symmetry.

\textbf{Evaluation Metrics.} The precision-recall metric with F-measure is used to evaluate the performance of symmetry detection, as introduced in \cite{14tsogkas2012learning}. To obtain the precision-recall curves, the detected symmetry response is first thresholded into a binary map, and then matched with the ground-truth symmetry masks. By changing the threshold value, the precision-recall curve is obtained and the best F-measure is computed.

\subsection{Results on Sym-PASCAL}
\vspace{-0.2em}
\subsubsection{SRN setting}\label{SRN-setting}
\vspace{-0.3em}
\begin{table}[t]
\begin{tabular}{ccc|c}
\hline
Architecture                  & Augumentation                                    & Conv1 & F-measure      \\ \hline
\multirow{4}{*}{shallow-deep} & \multirow{2}{*}{$1 \times$}                      & with  & 0.381          \\ \cline{3-4}
                              &                                                  & w/o   & 0.397          \\ \cline{2-4}
                              & \multirow{2}{*}{$0.8\times, 1\times, 1.2\times$} & with  & 0.371          \\ \cline{3-4}
                              &                                                  & w/o   & 0.396          \\ \hline
\multirow{4}{*}{deep-shallow} & \multirow{2}{*}{$1 \times$}                      & with  & \textbf{0.443} \\ \cline{3-4}
                              &                                                  & w/o   & \textbf{0.443} \\ \cline{2-4}
                              & \multirow{2}{*}{$0.8\times, 1\times, 1.2\times$} & with  & 0.384          \\ \cline{3-4}
                              &                                                  & w/o   & 0.397          \\ \hline
\end{tabular}
\vspace{-0.7em}
\caption{Performance of SRN under different settings on the Sym-PASCAL benchmark.}
\vspace{-1.6em}
\label{Tab-SRN-Setting}
\end{table}

SRN is first evaluated on the new benchmark with different settings, Tab.\ \ref{Tab-SRN-Setting}. \textbf{Architectures:} Tab.\ \ref{Tab-SRN-Setting} shows that SRN with the deep-to-shallow architecture (F-measure 0.443) performs significantly better than the shallow-to-deep architecture (F-measure 0.397). It confirms that the deep-to-shallow architecture is easier to reduce the residual than the shallow-to-deep one as the initialization is better.
\textbf{Data Augmentation:} Data augmentation can aggregate the training datasets. In this work, image rotation, flipping, up-sampling, and down-sampling (multi-scale) are used for data argumentation. For each scale, we rotate the training images every 90 degree and flip each one with different axis. The performance with/without multi-scale data argumentation is compared. Experiments show that the F-measure decreases with multi-scale augmentation, even though it produces more training data. The reason is analyzed as follows. The symmetry ground-truth is made up of curves with one-pixel thickness. The up-sampling operation produces curves that have thickness lager than one pixel, and the down-sampling operation produces discontinuous symmetry curves.
\textbf{Conv1:} FSDS \cite{04shen2016object} doesn’t use the conv1 stage of VGG as the size of receptive field is so small (only 5) that introduces local noise of symmetry (too small to capture any symmetry response). The negative impact of small receptive field with SRN is also observed. By pairwise comparison in Tab.\ \ref{Tab-SRN-Setting}, the F-measure without conv1 is slightly better than that with conv1.

\vspace{-0.6em}
\subsubsection{Performance Comparison}
\vspace{-0.4em}

Using the deep-to-shallow SRN with data augmentation but without conv1, we compare the performance of SRN with the state-of-the-art, as shown in Fig.\ \ref{figure8} and Tab.\ \ref{Tab-compare-sympascal}. All the compared results are generated by running the open source code with default parameter settings.

It's observed that the traditional methods perform poorly and are time consuming. The best F-measure of traditional methods is 0.174, indicating the challenge of the proposed benchmark. Lindeberg \cite{32lindeberg1998edge} runs fastest with 5.79s per frame. Levinshtein \cite{29levinshtein2009multiscale}, MIL \cite{14tsogkas2012learning}, Lee \cite{30sie2013detecting} and Particle Filter \cite{36widynski2014local} need much more running time for the complex features they used.

\begin{figure*}[t]
\begin{center}
\vspace{-0.3em}
\includegraphics[width=0.85\linewidth]{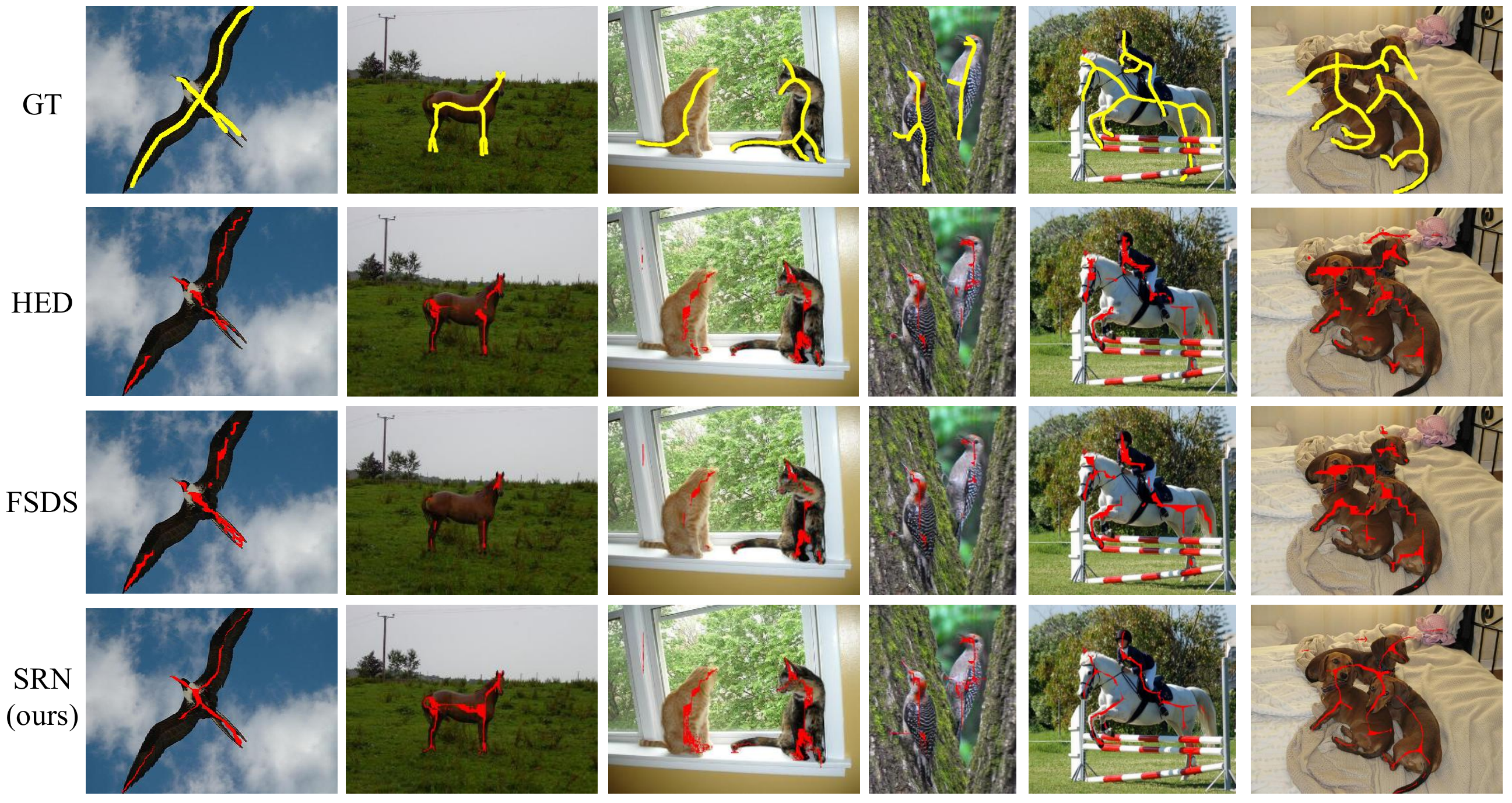}
\end{center}
\vspace{-2em}
\caption{Object symmetry detection results on the Sym-PASCAL dataset: the first and second columns for one-object images with/without complex background, the third and forth columns for multi-object images with/without complex background, and the last two columns for images with occluded objects. (Best viewed in color)} \label{figure9}
\vspace{-0.7em}
\end{figure*}

The end-to-end deep learning methods perform well. HED gets the F-measure 0.369 and uses only ten milliseconds to process an image. FSDS is degenerated to HED when the scale information is not used. Its F-measure reaches 0.418 when slicing and concatenating of each side-output is used. Our proposed SRN gets the best performance with F-measure 0.443 which outperforms the baseline HED approach by 7.4\%. It also outperforms the state-of-the-art method, FSDS, by 2.5\%.

To show the effectiveness of the end-to-end pipeline in complex backgrounds, we compare the proposed SRN with a two-stage approach composing of semantic segmentation/object detection and skeleton extraction. We choose the best segmentation network FCN-8s \cite{02long2015fully} to localize objects, and the skeleton method \cite{38DBLP:journals/pr/ShenBHWL11} to extract symmetry, getting F-measure 0.386, Fig.\ \ref{figure8}.
We also compare the FSDS \cite{04shen2016object} on the detection results from the state-of-the-art object detection methods, FasterRCNN \cite{DBLP:conf/nips/RenHGS15} and YOLO \cite{DBLP:journals/corr/RedmonDGF15}. As shown in Fig.\ \ref{figure8}, the F-measures are 0.343 and 0.354, respectively. Experiments results indicate that the proposed end-to-end learning approach is a more effective and efficient way to detect object symmetry than the two-stage approaches.

The object symmetry detection results by the state-of-the-art deep leaning approaches are illustrated in Fig.\ \ref{figure9}. From the first and second columns, it’s observed that the object symmetry obtained by our SRN approach in one-object images is more consistent with the ground-truth with/without complex background. The third and forth columns show examples that contain multiple objects, in which the proposed SRN approach achieves more accurate object symmetry detection results than other approaches. The last two columns show the results of images with occluded objects.

\begin{figure}[t]
\vspace{-0.5em}
\begin{center}
\includegraphics[width=0.75\linewidth]{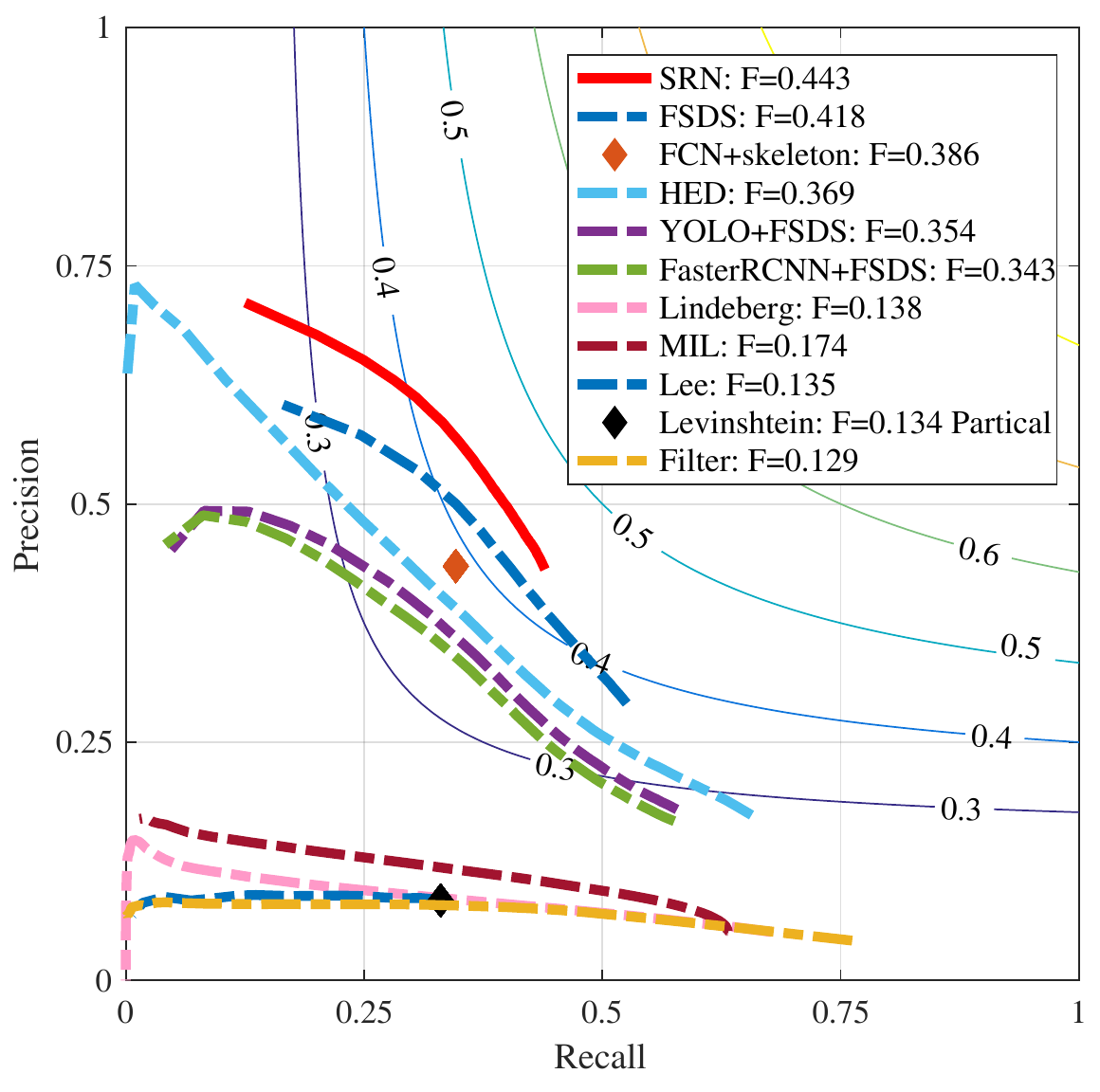}
\end{center}
\vspace{-2.5em}
\caption{Precision-recall comparison of different approaches on the Sym-PASCAL dataset.} \label{figure8}
\vspace{-0.5em}
\end{figure}
\begin{table}[t]
\small
\begin{tabular}{c|c|c}
\hline
Methods                      & F-measure & Runtime(s) \\ \hline
Partical Filter \cite{36widynski2014local}             & 0.129     & 25.30      \\
Levinshtein \cite{29levinshtein2009multiscale}                  & 0.134     & 183.87     \\
Lee \cite{30sie2013detecting}                        & 0.135     & 658.94     \\
Lindeberg \cite{32lindeberg1998edge}                   & 0.138     & 5.79       \\
MIL \cite{14tsogkas2012learning}                         & 0.174     & 80.35      \\ \hline
HED (baseline) \cite{03xie2015holistically}               & 0.369     & \textbf{0.10}$\dagger$      \\
FSDS \cite{04shen2016object}                        & 0.418     & 0.12$\dagger$       \\ \hline
FasterRCNN \cite{DBLP:conf/nips/RenHGS15}+FSDS \cite{04shen2016object} & 0.343 & 0.33$\dagger$ \\
YOLO \cite{DBLP:journals/corr/RedmonDGF15}+FSDS \cite{04shen2016object} & 0.354 & 0.12$\dagger$ \\
FCN \cite{02long2015fully}+\cite{38DBLP:journals/pr/ShenBHWL11} & 0.386     & 0.76$\dagger$      \\ \hline
SRN (ours)                   & \textbf{0.443}     & 0.12$\dagger$       \\ \hline
\end{tabular}
\vspace{-0.5em}
\caption{Performance comparison of the state-of-the-art approaches on the Sym-PASCAL dataset. $\dagger$GPU time with NVIDIA Tesla K80 }
\label{Tab-compare-sympascal}
\vspace{-3em}
\end{table}

\begin{figure*}[t]
\begin{subfigure}{0.3\textwidth}
\centering
\includegraphics[width=\linewidth]{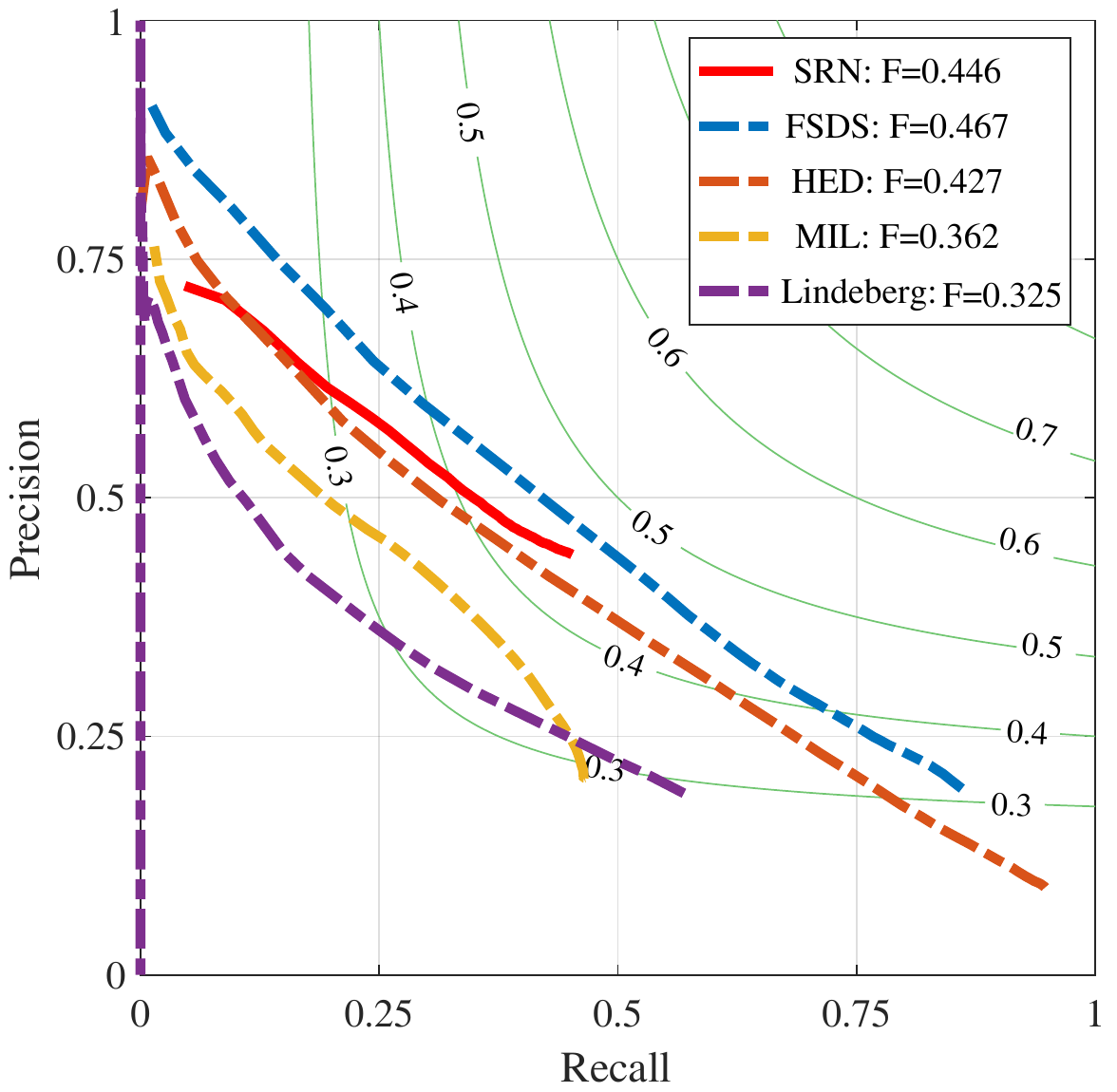}
\caption{SYMMAX} \label{figure11a}
\end{subfigure}
\hspace*{\fill} 
\begin{subfigure}{0.3\textwidth}
\centering
\includegraphics[width=\linewidth]{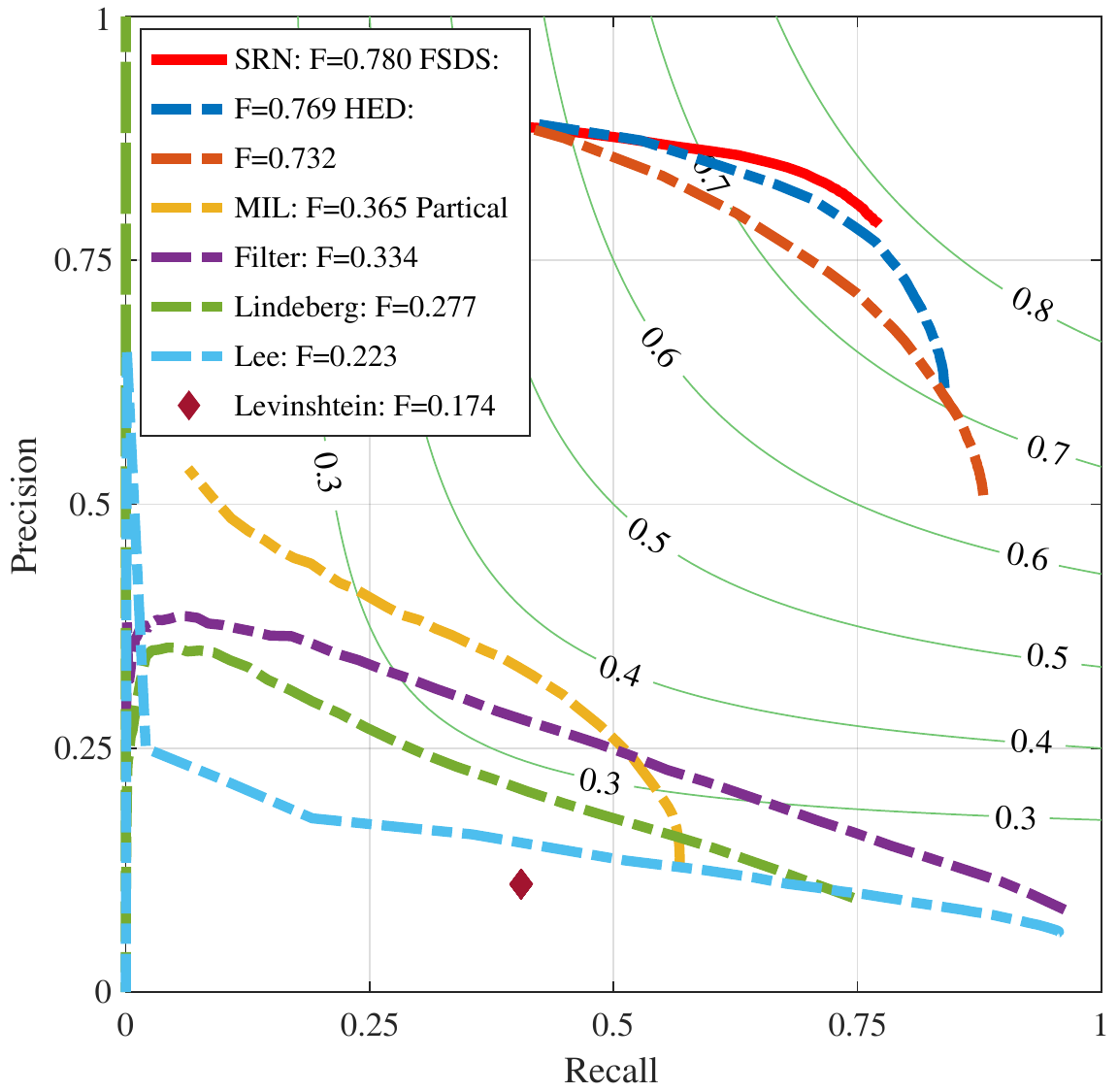}
\caption{WH-SYMMAX} \label{figure11b}
\end{subfigure}
\hspace*{\fill} 
\begin{subfigure}{0.3\textwidth}
\centering
\includegraphics[width=\linewidth]{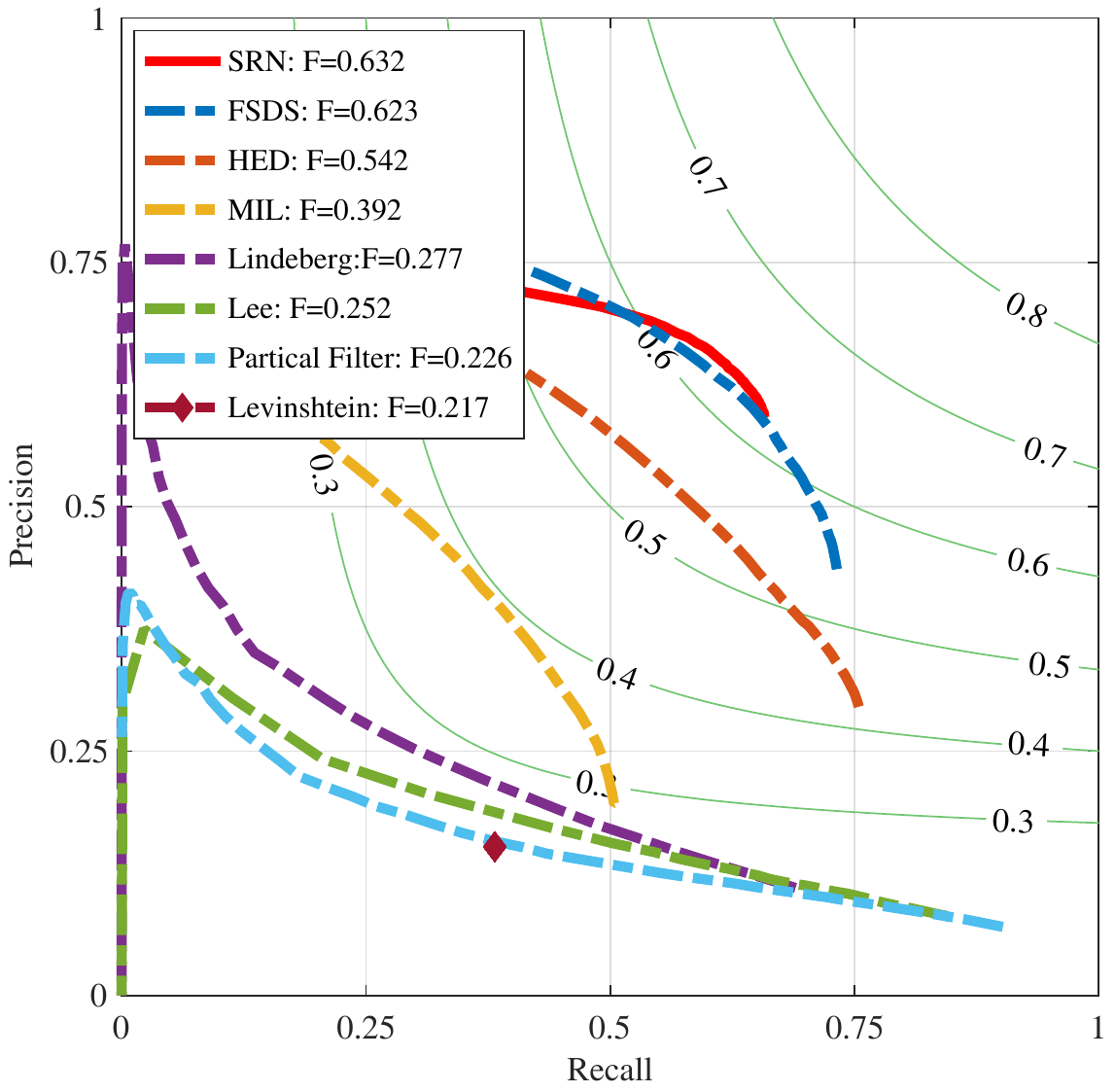}
\caption{SK506} \label{figure11c}
\end{subfigure}
\vspace{-0.6em}
\caption{The precision-recall curves of SYMMAX, WH-SYMMAX, and SK506 datasets.} \label{figure11}
\vspace{-0.6em}
\end{figure*}

\begin{table*}[t]
\centering
\begin{tabular}{c|ccccc|ccc}
\hline
\multirow{2}{*}{datasets} & Levinshtein & \multirow{2}{*}{Lee \cite{30sie2013detecting}} & Lindeberg & Particle         & \multirow{2}{*}{MIL \cite{14tsogkas2012learning}} & \multirow{2}{*}{HED \cite{03xie2015holistically} } & \multirow{2}{*}{FSDS \cite{04shen2016object}} & \multirow{2}{*}{SRN(ours)} \\
                          & \cite{29levinshtein2009multiscale}   &                                & \cite{32lindeberg1998edge} & Filter \cite{36widynski2014local} &                                &                                &                                 &                            \\ \hline
SYMMAX                    & --          & --                             & 0.360     & --               & 0.362                          & 0.427                          & \textbf{0.467}                  & 0.446                      \\
WH-SYMMAX                 & 0.174       & 0.223                          & 0.277     & 0.334            & 0.365                          & 0.732                          & 0.769                           & \textbf{0.780}             \\
SK506                     & 0.217       & 0.252                          & 0.227     & 0.226            & 0.392                          & 0.542                          & 0.623                           & \textbf{0.632}             \\
 \hline
\end{tabular}
\vspace{-0.7em}
\caption{Performance comparison of the state-of-the-art approaches on the SYMMAX, WH-SYMMAX, and SK506 dataset.}\label{Tab-compare-other}
\vspace{-1.5em}
\end{table*}

\begin{figure}[t]
\begin{center}
\includegraphics[width=\linewidth]{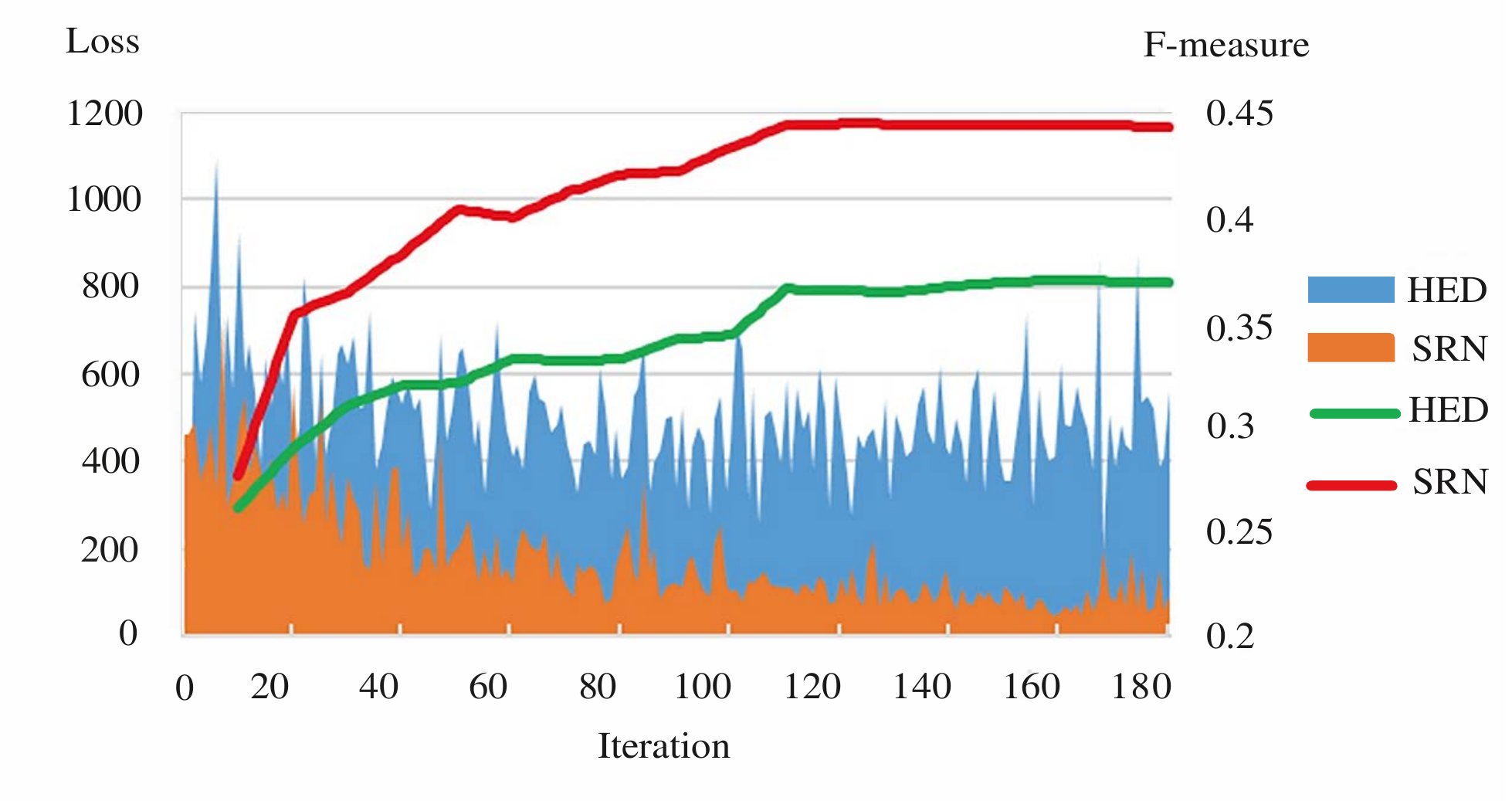}
\end{center}
\vspace{-2em}
\caption{The loss and F-measure comparison of HED and SRN. (Best viewed in color)} \label{figure10}
\vspace{-2em}
\end{figure}

\subsection{Results on Other Datasets}
\vspace{-0.5em}
The performances on other three symmetry datasets are shown in Fig.\ \ref{figure11} and Tab.\ \ref{Tab-compare-other}. Similar with Sym-PASCAL, the deep learning based methods get significantly better performance on all the datasets, especially for the simple image datasets, WH-SYMMAX and SK506. Compared with the baseline HED, the proposed SRN improves the F-measure from 0.427 to 0.446, 0.732 to 0.780, 0.542 to 0.632 on SYMMAX, WH-SYMMAX and SK506, respectively.

\vspace{-0.5em}
\subsection{Learning Convergence}
\vspace{-0.5em}

The learning convergence of the baseline HED and the proposed SRN is shown in Fig.\ \ref{figure10}. It can be clearly seen that HED has a problem of slow convergence during learning, despite the fact that it achieves good performance on the edge and symmetry detection tasks. The reason could be that the complex backgrounds of input images seriously interrupt the end-to-end (image-to-mask) learning procedure. Benefits from the output residual fitting, the loss curve of the proposed SRN tends to converge, Fig.\ \ref{figure10}. In addition, HED needs 12K learning iterations to get the best performance while SRN needs only 3K iterations to get the same performance.

\vspace{-1em}
\section{Conclusion}
\vspace{-0.5em}
Symmetry detection has great applicability in computer vision yet remains not being well solved, as indicated by the low performance (often lower than 50\%) of the state-of-the-art methods. In this work, we release a new object symmetry benchmark, as well as propose the Side-output Residual Network, establishing a strong baseline for object symmetry detection in the wild. The new benchmark, with challenges related to real-world images, is validated to be a good touchstone of various state-of-the-art approaches. The proposed Side-output Residual Network, with well-defined and stacked Residual Units, is validated to be more effective to perform symmetry detection in complex backgrounds. With the adaptability to object scales, the robustness to complex backgrounds, and the end-to-end learning architecture, the Side-output Residual Network has great potential to process a class of end-to-end (image-to-mask) computer vision tasks.

\vspace{-1em}
\section*{Acknowledgement}
\vspace{-0.5em}
This work is partially supported by NSFC under Grant 61671427, Beijing Municipal Science and Technology Commission under Grant Z161100001616005, and Science and Technology Innovation Foundation of Chinese Academy of Sciences under Grant CXJJ-16Q218. Tekes, Academy of Finland and Infotech Oulu are also gratefully acknowledged.

\vspace{-0.5em}
{\small
\bibliographystyle{ieee}
\bibliography{egbib}
}

\end{document}